\documentclass[10pt,letterpaper]{article}

\usepackage{cogsci}
\usepackage{apacite}

\usepackage{url}
\usepackage{tabularx}
\usepackage{times}
\usepackage{amsmath}
\usepackage{graphicx}
\usepackage{caption}
\usepackage{subcaption}
\usepackage[usenames, dvipsnames]{color} 
\usepackage{graphicx} 
\usepackage{svg}

\newcommand\ie{\emph{i.e.}}
\newcommand\eg{\emph{e.g.}}

\newcommand{\ignore}[1]{}

\newcommand{\Table}[1]{Table~\ref{#1}}

\newcommand{\Figure}[1]{Figure~\ref{#1}}

\usepackage{multirow}

\title{Simple Search Algorithms on Semantic Networks Learned from Language Use}

\author{{\large \bf Aida Nematzadeh}, {\large \bf Filip Miscevic}, and {\large
\bf Suzanne Stevenson} \\
Department of Computer Science \\
University of Toronto \\
\{aida,miscevic,suzanne\}@cs.toronto.edu
}

\begin{document}
\maketitle

\begin{abstract}
Recent empirical and modeling research has focused on the semantic fluency task because it is informative about semantic memory.
An interesting interplay arises between the richness of representations in semantic memory and the complexity of algorithms required to process it.
It has remained an open question whether representations of words and their relations \textit{learned from language use} can enable a simple search algorithm to mimic the observed behavior in the fluency task.
Here we show that it is plausible to learn rich representations from naturalistic data for which a very simple search algorithm (a random walk) can replicate the human patterns.
We suggest that explicitly structuring knowledge about words into a semantic network plays a crucial role in modeling human behavior in memory search and retrieval; moreover, this is the case across a range of semantic information sources.


\end{abstract}

\section{Introduction}

Semantic memory plays a significant role in cognition because it is the locus of storage for concepts and their relations.
There are a number of competing hypotheses for the representation of semantic memory, such as semantic networks \cite<\eg,>{collins.loftus.1975, steyvers.tenenbaum.2005}, vector space models \cite<\eg,>{landauer.dumais.1997}, and topic models \cite{griffiths.etal.2007}.
The content and structure of semantic memory is of great interest because it impacts how effectively people can store, search for, and retrieve information.

Recent work in computational modeling has illustrated in an interesting way the trade-off between the representation of semantic memory and the nature of the algorithms required to process it \cite{hills.etal.2012,abbott.etal.2015}.
The models in question focused on the semantic fluency task, in which people name as many members of a cue category as they can in a certain amount of time.
This task is informative about representation and processing of semantic memory because it requires people to access semantically-related words.
Based on their empirical data in such a task, \citeA{hills.etal.2012} argue that people follow an optimal foraging pattern that is similar to animals searching for food:
a semantic patch is \textit{exploited} until the rate of word retrieval is less than the long-term average rate of retrieval, and then a new patch of related words is \textit{explored}.

\citeA{hills.etal.2012} and \citeA{abbott.etal.2015} suggest that very different computational approaches are required to model this empirical behavior.
\citeA{hills.etal.2012} adopted a vector space representation of semantic memory -- one that encodes word--word co-occurrence patterns.
Using this representation, they showed that the best match to human behavior required a two-stage algorithm with an explicit strategy to switch from exploiting the current semantic patch to exploring a new patch.
In contrast, \citeA{abbott.etal.2015} showed that a simple random walk that operates uniformly was sufficient to model the pattern of behavior. To achieve this, their model used a semantic network representation that encoded relations among words from free association norms.
These results clearly demonstrate the interplay of representation and algorithm in replicating the same empirical data on semantic memory.

Having a semantic memory that is appropriately structured to support efficient real-time access might constitute a good balance in the representation/process trade-off.
But people must learn such a structure.
Creating a semantic network by directly encoding human association norms, as  \citeA{abbott.etal.2015} do, 
avoids the statistical learning problem that people face
\cite{jones.etal.2015}.
It has thus remained an open question whether representations of words and their relations \textit{learned from language use} can enable a simple search algorithm to mimic the observed behavior in the fluency task.

Our first contribution here is to show that this is indeed possible: we create a semantic network using learned meanings of words from a cognitively plausible computational model, and show that a simple, uniform random walk exhibits the observed foraging pattern of search. 
Moreover, we also show that if an explicit semantic network is created from the vector-space semantic information of \citeA{hills.etal.2012}, the same random walk algorithm on that network shows the desired match with human behavior.
We thus conclude that explicitly structuring knowledge about words into a semantic network plays a crucial role in modeling observed behavior in memory search and retrieval; moreover, this is the case across a range of semantic information sources (not solely in the case of free association data). 
We also perform structural analyses of the networks to consider the relation between their connectivity properties and their behavior.

\section{Semantic Fluency Data and Models}

\citeA{hills.etal.2012} argue 
that search through semantic memory is guided by the same strategy as that used by animals foraging for food.
In support of this view, they found that participant responses in a semantic fluency task (i.e., `name as many animals as you can in 3 minutes') came in bursts of semantically related ``patches'' (animal categories as defined by \citeA{troyer.etal.1997}, such as `pets' or `farm animals').  
Moreover, the timing of these responses was consistent with the marginal value theorem of optimal foraging in physical space \cite{charnov.1976}.
Specifically, the time it took for participants to retrieve the next novel item relative to the last one -- referred to as the inter-item retrieval time (IRT) -- increased with each item within a patch.
When the IRT exceeded the participant's average IRT across the entire trial, a switch into a different patch of semantically-related words occurred, and the IRT then decreased. This pattern can be seen in \Figure{fig:hills} in the Results section.

\citeA{hills.etal.2012} investigated the \ignore{goodness - F: ability a better word?}ability of different search algorithms to model this empirical data, using semantic representations of words learned by a vector space model, BEAGLE, on the Wikipedia corpus \cite{jones.etal.2007}.
They show that a two-stage algorithm best replicates the data, using local cues (word--word similarity) to find the next item within a patch, along with an explicit strategy to switch to a global cue (word frequency) to guide exploration of a new patch.
Moreover, they showed that a simpler search algorithm -- a random walk that used only the word--word similarities -- could not capture the observed foraging pattern.

In contrast, \citeA{abbott.etal.2015} showed that a simple random walk on a semantic network \textit{could} replicate human IRT patterns just as well as the two-stage algorithm of \citeA{hills.etal.2012}.
However, their semantic representation was created using human association norms \cite{nelson.etal.1998}.
\citeA{jones.etal.2015} raised the issue that
this semantic representation implicitly encodes the structure of a search process similar to the fluency task, thereby
making it possible for
a search algorithm simpler than that used by Hills et al.\
to replicate the empirical data.

In the remainder of the paper, we explore whether a structured representation that results in a simpler search and retrieval algorithm can be learned 
from the kind of data that people are naturally exposed to.
Similarly to \citeA{abbott.etal.2015} and in contrast to \citeA{hills.etal.2012}, we construct a semantic network to explicitly encode the appropriate relations among words.
However, unlike Abbott et al., our model learns these relations from a language corpus rather than simply encoding human association norms.
Moreover, unlike Hills et al., we use a corpus of child-directed speech to reflect more naturalistic language input, and use a semantic representation that explicitly draws on conceptual knowledge.

\section{Our Semantic Representation}

We briefly review our computational word learner,  then describe the process for constructing semantic networks.

\subsection{The Word Learner}
\label{sec:wd-lrner}

We use an incremental and probabilistic cross-situational learner shown to
mimic a range of child and adult behaviors in vocabulary learning
\cite{fazly.etal.2010.csj}.
The model takes as input a sequence of utterance--scene pairs, $U$--$S$, where
$U$ represents the linguistic input to a child, and $S$ represents the
non-linguistic data a child perceives in language learning.
The input is highly ambiguous, as the mapping between individual words in $U$
and the relevant semantics in $S$ is not explicitly indicated -- $U$ is
represented as a set of words, and $S$ as a set of semantic features: 
\begin{table}[h!]
\small{
\vspace{-0.05cm}
\begin{tabular}{l}
{\bf $U$:} $\{$\emph{crocodile}, \emph{float}, \emph{in}, \emph{the}, \emph{river}$\}$ \\
{\bf $S$:} $\{$ \dots, \textsc{reptile}, \textsc{vertebrate}, \dots,
\textsc{body-of-water}, \dots $\}$
\end{tabular}
}
\vspace{-0.25cm}
\end{table}

\noindent
From such input, the model uses an incremental version of
expectation-maximization to learn a probability distribution $P(.|w)$ for each
word $w$ over all observed features.

The utterances in the input are taken from a corpus of child-directed speech.
To create the associated scene representations, each word in the corpus is
entered into a gold-standard lexicon. (This lexicon is never seen by the
model.)
Each word in the lexicon has a set of semantic features representing its gold-standard meaning.
The features for each animal word (and nouns in general) are the names of each ancestor node (hypernym) of the word's first sense in WordNet\footnote{\url{http://wordnet.princeton.edu}}.
Each noun is thus represented by definitional features that reflect conceptual knowledge: general features such as {\sc object}, which appear with many words, and more specific features such as {\sc reptile}, which appear with fewer words.  For example:
\begin{center}
\begin{tabular}{ l }
{\em crocodile}: \{ {\sc crocodilian reptile}, {\sc diapsid}, {\sc reptile}, \\ 
\hspace{01.7cm}  {\sc vertebrate}, $\cdots$, {\sc whole}, {\sc object}, $\cdots$ \} \\
\end{tabular}
\end{center}
\vspace{-0.1cm}

\noindent
Scene $S$ for utterance $U$ is formed by taking the union of the gold-standard semantic features for all words in $U$. 
Thus the semantic input to the model represents naturalistic features that are distributed realistically across related entities, and reflect a conceptual hierarchy intended to approximate the type of conceptual categories children are forming.

An interesting property of the learner is that the learned meaning probabilities for a word $w$, $P(f|w)$ for observed features $f$, reflects not only the co-occurrences of $w$ with its gold-standard features:  The probabilities importantly capture the influence of contextual features in the input as well.
For example, \textit{crocodile} and \textit{hippopotamus} will be distinguished by high probabilities for the definitional features $P(\textsc{reptile}|\textit{crocodile})$ and $P(\textsc{mammal}|\textit{hippopotamus})$, but are both likely to have a higher than chance value for the feature \textsc{body-of-water} since both animals live in rivers.
Thus the learned semantic representation in the model captures both definitional and contextual similarities of words.

\subsection{Constructing a Semantic Network}
\label{sec:const-nets}

Other recent research has used free-association norms or conceptual hierarchies like WordNet as the basis for a semantic network; two words are connected by an edge in the network if there is a direct connection between them in the representation \cite{steyvers.tenenbaum.2005,abbott.etal.2015}.  
By contrast, for our meaning representation (as for BEAGLE data), the appropriate network connections among the words must be determined by considering \textit{how related} any pair of words is in that representation (since all words are implicitly more or less related).
We follow \citeA{nematzadeh.etal.2014.cogsci} in their approach to creating a semantic network over our model's learned meaning representations.  
Since we aim to model the empirical data from \citeA{hills.etal.2012} that looked at semantic fluency in the category of animals, we focus on the subset of words in our training data that occur in the dataset of \citeA{hills.etal.2012}.
Each such word is represented as a node in the network, and pairs of nodes are connected if the cosine similarity of their associated meaning probability vectors exceeds a certain threshold $\tau$.
The meaning similarity serves as the weight on an inserted edge.

We experiment with various values for the edge-threshold $\tau$, and at higher values, 
the resulting network becomes somewhat disconnected: groupings of very similar words form sets of connected components, usually animals of a similar subcategory (e.g. `farm animals' or `pets'). 
This reflects the fine-grained differences in word meaning that the learner has acquired. 
Because these learned representations do not completely capture taxonomic knowledge -- i.e., that   `animal' is a subsuming category of those groupings naturally occurring in the network -- we treat the word \textit{animal} differently in deciding on its network connections.\footnote{The calculation of model probabilities $P(f|w)$ entails that general features like \textsc{animal} (that many words share) have lower probability than specific features that distinguish the words.}
Specifically, we use a lower threshold, $\tau{_a}$, to determine when to add edges including the node for  \textit{animal}.
This ensures that \textit{animal} is connected to a number of the groupings of animals, and increases the connectivity of the network.
(Future work will look at mechanisms as in \shortciteA{nematzadeh.etal.2015.emnlp} for adequately capturing the meanings of hierarchically organized entities.)

The resulting graph may not be fully connected.
Since the fluency task starts with the cue word \textit{animal} and can only reach nodes that are directly or indirectly connected to it, we take the semantic network for our purposes to be the connected component of the graph that includes  \textit{animal}.
The number of nodes in the semantic network may thus be smaller than the number of observed animal words.

\section{Experimental Methods}

\subsection{The Semantic Networks}

The child-directed speech that forms the basis for the input to our word learner is the Manchester corpus \cite{theakston.etal.2001} of CHILDES \cite{macwhinney.2000}.
Of the $518$ unique animals classified by \citeA{hills.etal.2012} using the categories described by \citeA{troyer.etal.1997}, $111$ of these are present in the full corpus and thus in our gold standard lexicon.  However, only $93$ of these appear in the $481$K-word corpus ($120$K utterances) we use for training.
Thus, a semantic network of learned meanings -- called a Learner network -- will have a maximum of $94$ nodes ($93$ words from the animal subcategories plus \textit{animal} itself).

Recall that the learned representations from our model reflect both definitional and contextual aspects of word meaning; this contextualization of meaning has been shown to influence the structure of resulting semantic networks \shortcite{nematzadeh.etal.2014.cogsci}.
For comparison, we create ``gold-standard'' semantic networks, called Gold, whose edge connections are determined using the gold-standard (definitional) meanings rather than the learned meanings.
These networks enable us to see the impact of having the hierarchical semantics from WordNet without the contextually learned aspects of meaning.
The Gold networks have a maximum of $112$ nodes ($111$ animal terms+\textit{animal}).

Finally, we used the same method to create BEAGLE semantic networks using the data reported by \citeA{hills.etal.2012}.
BEAGLE consists of word co-occurrence data that encodes contextualized meanings; however, some  hierarchical conceptual knowledge is reflected in the $400$M-word Wikipedia corpus it was trained on.
The BEAGLE data contains 364 animal words that appear in \citeA{hills.etal.2012}, and thus these networks have a maximum of 365 nodes.

For all semantic networks, we use cosine similarity as the (potential) edge weights, and consider various levels of the thresholds $\tau{_a}$ (for edges that include the word \textit{animal}) and $\tau$ (for all other edges) on these weights for inclusion of edges.\footnote{Our code and data are available at \url{https://github.com/FilipMiscevic/random_walk.git}.}

\subsection{Simulating Behavior with Random Walks}

Our goal is to see whether the structure of our semantic networks is sufficient to obtain the observed foraging behavior using a simple, uniform search algorithm.
To that end, we perform random walks with variations as discussed by \citeA{abbott.etal.2015}.
Each random walk begins at the word \textit{animal} to simulate the fact that \textit{animal} is the cue for the fluency task (\ie, ``name as many animals as you can'').
Each step in a random walk -- \ie, the move from the current node $n_c$ to the next node $n_n$ -- is determined by a probabilistic selection over the edges incident on $n_c$.
The selection process may choose the edge to follow in proportion to the edge weights (a weighted walk), or use a uniform distribution over all edges connected to $n_c$ (an unweighted walk).
(A further variation in which there is a probability $p$ of jumping back to the word \textit{animal} after any step in the random walk had no appreciable impact on our results, so we do not report that method here.)
Due to the probabilistic nature of the algorithm (in selecting edges to traverse), we report results averaged over $282$ random walks for each network under parameter settings of interest. 

To reflect the time limit in the semantic fluency task, 
\citeA{abbott.etal.2015} fix the number of steps in the random walks to produce approximately the same number of words as human participants.
Because this walk length 
is dependent on properties of the graph being traversed, Abbott et al.\ set this for each network, using walk lengths of $45$ with the BEAGLE data and $2000$ on their own semantic network.
We take an alternative approach: Instead of picking one walk length to produce a certain number of words, we explore the interaction of different walk lengths with parameters of the networks to see which combinations lead to an appropriate number of words produced.
We aim for a range of number of words produced around that of people -- i.e., $37 \pm 5$.

\subsection{Evaluating IRTs and Patch Switches}

In assessing the fit of the random walks to human data, we use the same mapping of steps in the walk to the IRT as used by \citeA{abbott.etal.2015}. Only the first visit to a node counts as producing a word (just as repeats of words are not counted in the human task); the IRT is thus counted between such first visits: i.e., the IRT is the number of steps in the walk between a node $n_i$ the first time it is visited and the next node $n_j$ in the walk that has not been previously visited.  Any nodes revisited between such an $n_i$ and $n_j$ increase the IRT between them.

Patch switches occur in the fluency task when participants switch from listing animals in one subcategory (such as `farm animals') to another (such as `pets').
Motivated by findings in \citeA{hills.etal.2009b}, we use a ``fluid patch model'' with the \citeA{troyer.etal.1997} categories of animals in analyzing our results.
This approach takes into account that animals may belong to multiple categories: a patch switch is considered to have occurred whenever the current novel word and the next novel word in the walk do not have \textit{some} category in common.
Patch switches are used to determine the patch entry position in analyzing the match of the random walks to human data (e.g., ``1'' represents the first item in a patch; ``-1'' is the last word before a patch switch; see \Figure{fig:allIRTs}).

To assess whether our networks match the human IRT pattern, we consider specific thresholds for the ratio of the IRT at certain points to the overall mean IRT of the random walk.
For the patch entry point ($1$), this ratio for the human data is around $1.2$ (cf.~\Figure{fig:hills}); we consider a minimum threshold $1.1$ as achieving a fit, with a stricter ratio of $1.2$ indicating a better match to human data.\footnote{\citeA{hills.etal.2012} note that to mimic foraging the value simply needs to be higher than the average IRT.}
For the IRT at position $2$, where there is a decrease following the patch switch, we similarly set a maximum threshold of $0.80$ of the mean IRT over all.
For all other positions, the ratio of IRT to mean IRT must be less than or equal to $1.0$.
We report walks as matching human data when they meet all these thresholds (and note when the stricter of the patch entry thresholds is met).

\section{Experimental Results}

\subsection{Parameter Search and Selection}

Several parameters influence both the number of words produced in a random walk on our networks, and the precise pattern of IRTs and patch switches.  The thresholds $\tau$ and $\tau{_a}$ used in determining the edges to include in the networks (for non-\textit{animal} and \textit{animal} nodes, respectively) affect both how connected the network is and the actual pattern of connectivity (e.g., all over loosely connected, or a disjoint set of connected components).  For example, having fewer edges does not necessarily lead to less connectivity, but might increase the path length between a given pair of words.

Similarly, the number of steps the random walk is allowed -- the ``walk length'' $L$ -- clearly influences the number of words produced, but it affects the patterning as well.  For example, longer walks have more opportunity to explore more subcategories of words, which can affect the patch switching.  Also, a longer walk does not necessarily mean that more words are produced -- it might simply raise the IRTs by spending more time revisiting nodes.

Given that the structure of the network and the random walk length interact to produce both a certain number of words and a given IRT pattern, we perform a parameter search over pairs of reasonable values of the edge threshold $\tau$ and the walk length $L$.  
(We fix $\tau{_a}$ for connecting \textit{animal} at $0.40$, which we found to give good results across all networks.)
We vary $\tau$ in increments of $0.05$, with the range chosen for each type of network based on preliminary experimentation. We vary $L$ from $35$ (the approximate number of words produced by people) to $135$ (within which all networks showed humanlike behavior for some value of $\tau$).

We search over $\tau \times L$ to find the combinations that yield walks over Learner, Gold, and BEAGLE that match the human pattern of responses. In particular, we looked for parameters that: (i) produce a range of number of words similar to that of people, and (ii) produce an IRT pattern that matches that of people (as detailed in Methods).  Instead of simply finding one parameter combination that achieves these goals, we consider the number of such walks across a range of parameter settings to indicate the robustness of an approach to semantic representation.

\subsection{Overall Patterns Observed}

Generally $\tau$ and $L$ work together to produce the desired output patterns -- i.e., the higher $\tau$'s need higher $L$'s to produce the right number of words.
We select a range of four $\tau$ values (Learner [$.70$--$.85$], Gold [$.75$--$.90$], BEAGLE [$.40$--$.55$]) and nine settings of $L$ ($60$--$100$) that exhibit the best performance in showing human behavior (as in (i) and (ii) above).  This yields a set of $36$ walks in each of the weighted and unweighted settings to analyze; see \Table{tab:walksummary}.
\begin{table}
\begin{center}
\small{
\begin{tabular}{| c |r |r| r | r |r| r| }  
\cline{2-7} 
\multicolumn{1}{c}{} & \multicolumn{3}{|c|}{Weighted} &\multicolumn{3}{|c|}{Unweighted} \\
\hline
Network & N & IRT & IRT+ & N & IRT & IRT+  \\
\hline
Learner &  81 & 44 & 22  & 92 & \textbf{64} & 11\\
 Gold   &  56 & 14 & 14  & 56 & \textbf{33} & 19 \\
BEAGLE  &  25 & \textbf{19} & 0    & 25  & 11 & 0 \\ 
\hline
\end{tabular}
}
\end{center}
\vspace{-0.15cm}
\caption{\small{The \textit{percentage} of $36$ walks (weighted and unweighted), varying $\tau \times L$, that match people with respect to (a) the number of words produced (N); (b) N and the IRT pattern (IRT); (c) IRT at the stricter threshold on patch entries (IRT+).}}
\label{tab:walksummary}
\vspace{-0.4cm}
\end{table}

Overall, BEAGLE performs somewhat better with weighted walks and our networks somewhat better with unweighted walks.
The high $\tau$ in our networks means edge weights have a small range and are thus very similar -- i.e., they are not much more informative than picking uniformly.
Also, BEAGLE is trained on a corpus over $800$ times the size of ours, so our Learner weights may simply be noisier. 

We find that the best performance for BEAGLE (weighted) only matches the target human pattern for $19\%$ of the walks; the best for Gold does so for $33\%$ and the Learner for $64\%$ (both unweighted). 
Even with weighted walks, our Learner achieves the pattern in $44\%$ of the walks, over twice the number of BEAGLE.  
We believe that our learned representations, which encode both conceptual knowledge from WordNet coupled with contextual influences from corpus co-occurrences, more robustly reflect the nature of the similarity relations among words for this task. 
Thus, Learner performs better than both Gold and BEAGLE that each only (primarily) capture one of these types of knowledge.

Interestingly, we get these patterns with walk lengths in the range of $60$--$100$, where \citeA{abbott.etal.2015} used lengths of $2000$ to produce words at the rate of people.
Perhaps word co-occurrence data more directly captures relations amongst a wide variety of words compared to the association norms of their data.
Future analysis of their network compared to ours may reveal why their walks apparently revisit nodes much more frequently.

\subsection{Comparing Best Results}

\begin{figure}
\begin{subfigure}{.25\textwidth}
  \centering
  \includegraphics[trim=0 0 10 47, clip,scale=0.35]{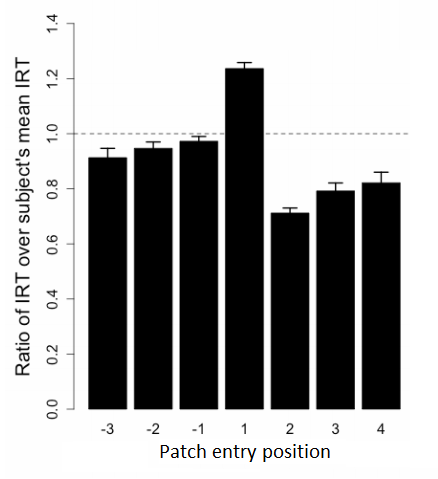}
  \caption{Human data }
  \label{fig:hills}
\end{subfigure}%
\begin{subfigure}{.25\textwidth}
  \centering
  \includegraphics[trim=0 0 0 43, clip,scale=0.25]{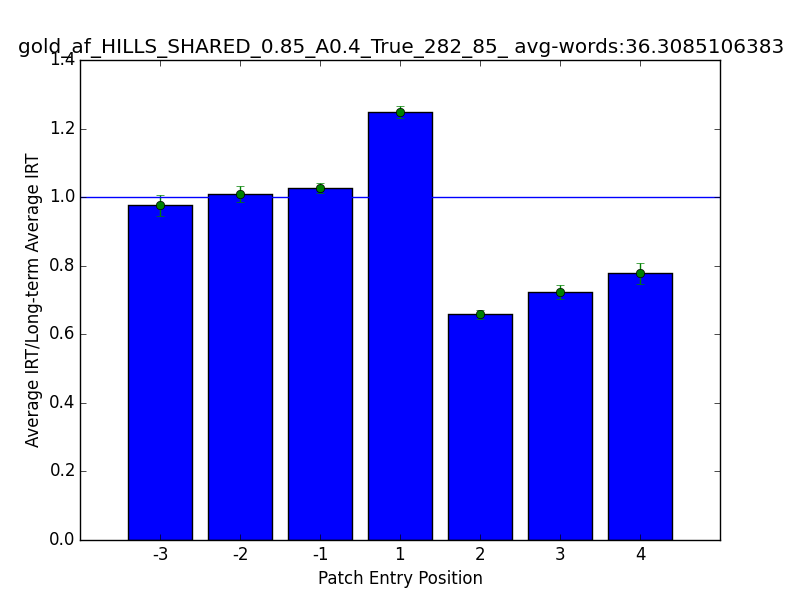}
  \caption{Gold network}
  \label{fig:sfig2}
\end{subfigure}
\begin{subfigure}{.25\textwidth}
  \centering
  \includegraphics[trim=0 0 0 58, clip,scale=0.25]{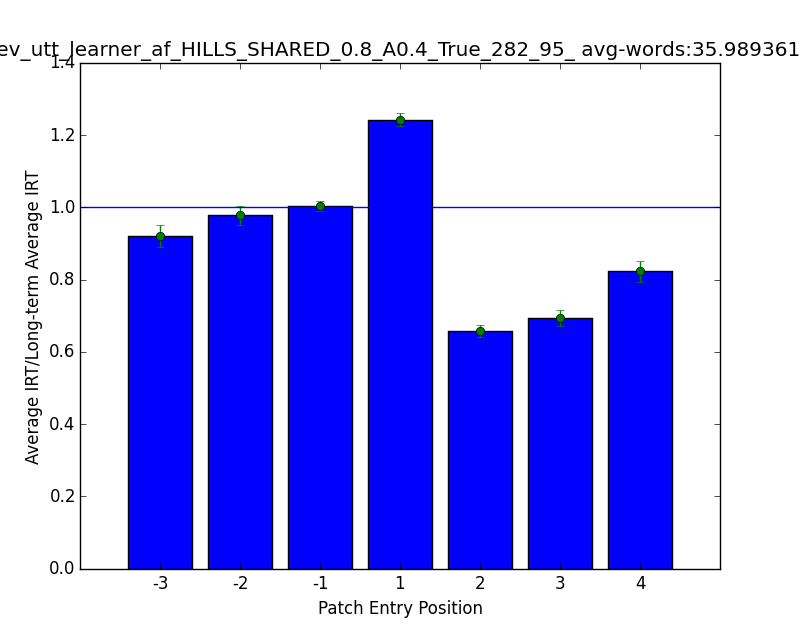}
  \caption{Learner network}
  \label{fig:sfig3}
\end{subfigure}%
\begin{subfigure}{.25\textwidth}
  \centering
  \includegraphics[trim=0 0 0 43, clip,scale=0.25]{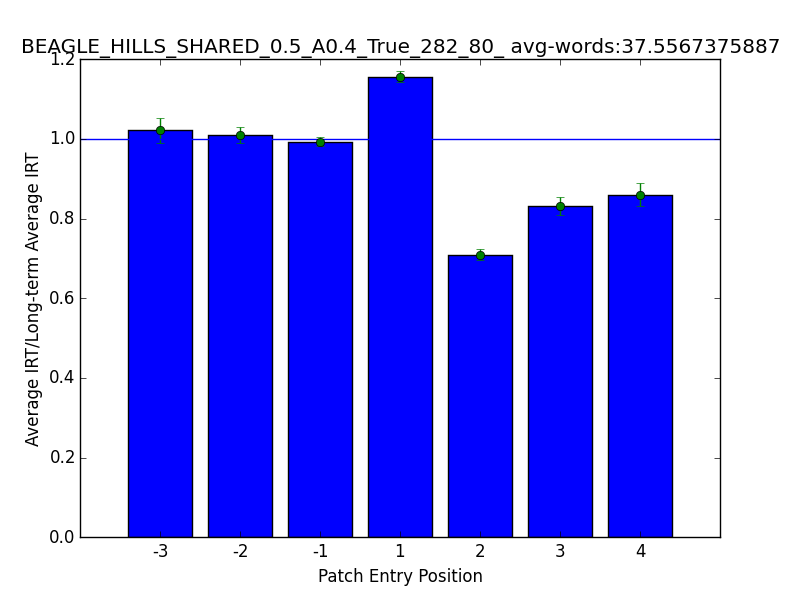}
  \caption{BEAGLE network}
  \label{fig:sfig4}
\end{subfigure}
\caption{\small{(a) Human IRTs reproduced from \citeA{hills.etal.2012}.\\(b--d) Modeling IRTs in  weighted random walks using the parameters described in Comparing Best Results.}} 
\label{fig:allIRTs}
\end{figure}

To look more closely at specific patterns, we compared the networks under the best $\tau$ parameter for each (Learner: 0.80, Gold: 0.85, BEAGLE: 0.50), with the full range of $L=35-135$; see \Table{tab:bestwalksummary}. 
For these settings, we found all networks did the same or slightly better using a weighted walk compared to unweighted. 
All networks perform very similarly, with the primary difference that the Learner network matches the human target behavior in more walks.
Moreover, both Gold and Learner meet the stricter IRT ratio of 1.2 in most cases of weighted walks, while BEAGLE only meets the less strict ratio of 1.1.
See \Figure{fig:allIRTs} for the results of a sample walk ($L =$ $95$ [Learner], $85$ [Gold], $80$ [BEAGLE]).

\begin{table}
\begin{center}
\small{
\begin{tabular}{ |c |r |r| r | r |r | r| }  
\cline{2-7}
\multicolumn{1}{c}{} & \multicolumn{3}{|c|}{Weighted} &\multicolumn{3}{|c|}{Unweighted} \\
\hline
  Network & N & IRT & IRT+ & N & IRT & IRT+  \\
\hline
 Learner & 52 & \textbf{38} & \textbf{38}  & 38 & 33 & 14 \\
 Gold    & 33 & \textbf{29} & \textbf{29}  & 33 & \textbf{29} & \textbf{29} \\
 BEAGLE &  33 & \textbf{29} & 0     & 33 & 14 &  0 \\ 
\hline
\end{tabular}
}
\end{center}
\vspace{-0.15cm}
\caption{\small{The \textit{percentage} of $21$ walks (weighted and unweighted), for best $\tau$ per network, that match people with respect to (a) the number of words produced (N); (b) N and the IRT pattern (IRT); (c)~IRT at the stricter threshold on patch entries (IRT+).}}
\label{tab:bestwalksummary}
\vspace{-0.4cm}
\end{table}

In summary, human-like IRT patterns were observed for random walks on each of the three networks.
Importantly, this includes random walks using the BEAGLE data, which \citeA{hills.etal.2012} previously showed could not produce such a pattern when used directly.
This demonstrates that creating a semantic network from the BEAGLE representation imposes important structure on the raw co-occurrence data, helping the network to focus on meaningful word--word connections.
Moreover, the fact that our Learner network shows a very good match to human behavior demonstrates that appropriate representations for a semantic network can be acquired by a cognitively-plausible word learner.

\subsection{Analyzing the Structure and Semantics of Networks}

Previous research suggests that a \textit{small-world} network -- a sparse graph with highly-connected sub-networks organized around ``hubs'' -- enables efficient access to semantic information \cite{steyvers.tenenbaum.2005}.
The idea is similar to foraging: first the 
hubs are explored, and then a new sub-network connecting to the matched hub is exploited.
Indeed, a semantic network created from the association norms used by \shortciteA{abbott.etal.2015} has been shown to have a small-world structure \shortcite{steyvers.tenenbaum.2005}.

As in \shortciteA{nematzadeh.etal.2014.cogsci}, we calculate a ``small-worldness'' score ($\sigma$)  for each of our networks, using well-known graph metrics; when $\sigma > 1$, the network conforms to a small-world structure. See \Table{tab:sw-fscore} for the best networks as in \Figure{fig:allIRTs}.
We find that all the networks 
that exhibit the target IRT pattern have a small-world structure; in other words, a small-world structure may be \textit{necessary} in producing the human pattern.
However, having a small-world structure is not \textit{sufficient}: most of the networks under the wide range of parameter settings we examined have small-world structure, but not all exhibit the foraging behavior.

\begin{table}
\begin{center}
\small{
\begin{tabular}{ |l |r |r| r | r |r | r| }  
\cline{2-7} 
\multicolumn{1}{c}{}  & \multicolumn{3}{|c|}{Structure} &\multicolumn{3}{|c|}{Semantics} \\
\hline
  Network & $\sigma$ & Nodes & Edges & P & R & F-score  \\
\hline
 Learner&  24 & 88   & 205  & 0.75 & 0.49 & 0.59\\
 Gold   &  24 & 112  & 302  & 0.72 & 0.50 & 0.59\\
 BEAGLE &   7 & 136  & 304  & -    &  -   &  - \\ 
\hline
\end{tabular}
}
\end{center}
\vspace{-0.3cm}
\caption{\small{The small-world and clustering results for best networks. $\sigma$ is small-worldness; P and R are average precision and recall, respectively.}}
\label{tab:sw-fscore}
\vspace{-0.5cm}
\end{table}

We observe that an appropriate \textit{graph structure} on its own cannot guarantee efficient search and retrieval: For that, the content of the sub-networks need to appropriately link \textit{semantically-related} words.
Indeed, \shortciteA{abbott.etal.2015} also find that their network captures appropriate groupings of animals.
We considered whether our networks also reflect the structure of animal subcategories.
For the Learner and Gold networks, we can do this by removing the \textit{animal} node and its edges (which we added as the cue word for the random walk), and then labeling each connected component of the network with the most frequently occurring category from  \citeA{troyer.etal.1997}.
We take a mean of precision and recall for each such cluster, weighted by its size, and compute the F-score (see \Table{tab:sw-fscore}).
Although not all subcategories of animals are connected to each other (lower recall), the sub-networks have mostly animals from the same subcategory (high average precision), supporting the observed human-like patch switching.

Unfortunately, the networks from BEAGLE do not form such connected components, making this approach to clustering analysis inappropriate.  
We note here that \shortciteA{abbott.etal.2015} claim the BEAGLE data shows only a ``weak signature of animal clusters''.
We also observe that the small-worldness value is overall larger in our networks than that of BEAGLE; these properties of BEAGLE networks may explain why they do not perform as robustly as our networks in replicating the behavioral data.

\section{Discussion and Future Work}

There is an interesting interplay between the richness of representations in semantic memory and the complexity of algorithms required to process it. 
We show that it is plausible to learn rich representations from naturalistic data for which a very simple search algorithm (a random walk) is enough to replicate the patterns observed in people.
Two key factors play a role in the success of our approach: (1)~Our learned representations capture the hierarchical relations among words as well as their contextual similarities. 
(2)~We explicitly impose a structure onto our learned representations by creating a semantic network in which words are connected only if their similarity exceeds a certain threshold.

Our work builds on recent research by \shortciteA{hills.etal.2012} and \shortciteA{abbott.etal.2015} in which different representation--algorithm pairs (vectors of co-occurrence statistics and strategic search vs.\ association norms and random search) replicate the same behavioral data from a fluency task: people name animal words from a  
subcategory (\eg, pets) until their rate of retrieval is less than the long-term average rate of retrieval, and then they switch to a new subcategory (\eg, farm animals).
Importantly, our approach has the advantage that our representations  
are learned from naturalistic language learning data.
Although here we created the semantic networks using the final learned representations of the model, 
these networks can also be acquired incrementally during word learning \shortcite{nematzadeh.etal.2014.emnlp}.

We further demonstrate 
that a random walk on a semantic network created from the vector representations of \shortciteA{hills.etal.2012} can produce the observed human pattern. This shows that the co-occurrence statistics learned from a large corpus encodes the required semantic information; however, the explicit structure of a semantic network is needed to simplify the search process. 
Moreover, our analysis reveals that to replicate the behavioral data all semantic networks (using the various representations) need to have certain connectivity properties -- \ie, they consist of highly-connected components, and most nodes are reachable from other nodes via relatively short paths.

\setlength{\bibleftmargin}{.125in}
\setlength{\bibindent}{-\bibleftmargin}
\bibliographystyle{apacite}
\bibliography{nematzadeh,sw}

\end{document}